\documentclass[letterpaper,journal]{IEEEtran}

\usepackage{amsmath,amsfonts}
\usepackage{algorithmic}
\usepackage{array}
\usepackage[caption=false,font=normalsize,labelfont=sf,textfont=sf]{subfig}
\usepackage{textcomp}
\usepackage{stfloats}
\usepackage{url}
\usepackage{verbatim}
\usepackage{graphicx}
\usepackage{cite}
\usepackage{etoolbox}
\usepackage{multirow}
\usepackage[ruled,vlined]{algorithm2e}
\usepackage[T1]{fontenc}

\usepackage{xcolor}

\begin{document}

\title{Enhancing Drone Light Shows Performances: Optimal Allocation and Trajectories for Swarm Drone Formations}

\author{\IEEEauthorblockN{Yunes ALQUDSI}\\
\IEEEauthorblockA{\textit{\textsuperscript{1}Aerospace Engineering Department, Faculty of Aeronautics and Astronautics, Kocaeli University, Kocaeli, Turkiye}\\
\textit{\textsuperscript{2}Interdisciplinary Research Centre for Aviation and Space Exploration (IRC-ASE), KFUPM, Dhahran, Saudi Arabia}\\
Email: yunes.alqadasi@kocaeli.edu.tr}
}

\maketitle

\begin{abstract}
Drone light shows (DLShows) represent a rapidly growing application of swarm robotics, creating captivating aerial displays through the synchronized flight of hundreds or thousands of unmanned aerial vehicles (UAVs) as environmentally friendly and reusable alternatives to traditional pyrotechnics. This domain presents unique challenges in optimally assigning drones to visual waypoints and generating smooth, collision-free trajectories at a very large scale. This article introduces the Unified Assignment and Trajectory Generation (UATG) framework. 
The proposed approach concurrently solves two core problems: the optimal assignment of drones to designated goal locations and the generation of dynamically feasible, collision-free, time-parameterized trajectories. The UATG framework is specifically designed for DLShows, ensuring minimal transition times between formations and guaranteeing inter-drone collision avoidance. A key innovation is its exceptional computational efficiency, enabling the coordination of large-scale in real-time; for instance, it computes the optimal assignment and trajectories for 1008 drones in approximately one second on a standard laptop.
Extensive simulations in realistic environments validate the framework's performance, demonstrating its capability to orchestrate complex formations—from alphanumeric characters to intricate 3D shapes—with precision and visual smoothness.
This work provides a critical advancement for the DLShow industry, offering a practical and scalable solution for generating complex aerial choreography and establishing a valuable benchmark for ground control station software designed for the efficient coordination of multiple UAVs. A supplemental animated simulation of this work is available at \url{https://youtu.be/-Fjrhw03594}.
\end{abstract}


\section{Introduction}
In recent years, drones have experienced widespread adoption across various industries, thanks to technological advancements and supportive drone policies \cite{yucesoy2025role}. This has opened up avenues for exploring innovative applications and creative uses of drones \cite{yumnam2025utilising, alqudsi2025advancements, aboelezz2019wind}. The availability of affordable Unmanned Aerial Vehicles (UAVs) like multirotor drones, coupled with advancements in Artificial Intelligence (AI) and control theory, has facilitated the utilization of coordinated swarms of UAVs \cite{liu2022rapid, alqudsi2024swarm}. One particularly captivating application lies in the utilization of drone swarms for pattern formation control, whereby synchronized aerial performances captivate audiences.

The rise of cost-effective UAVs has introduced exciting opportunities across diverse fields such as surveillance \cite{speth2022deep}, search and rescue \cite{skinner2018uav}, aerial inspection \cite{alharbi2025ai}, landmine detection \cite{colorado2017integrated}, defense \cite{boyle2020drone}, and beyond \cite{mohanty2023drone, alqudsi2025towards}. Such applications require robust, flexible, and scalable coordination frameworks specifically developed for UAV swarms. By leveraging coordination and local interactions among UAVs and their environment, swarms of robots can perform complex tasks that surpass the capabilities of a single agent \cite{ranjan2024drone}. By forming a swarm of aerial robots, agents establish communication and coordination channels, enabling them to leverage the resources and capabilities of their fellow agents, improving individual performance and enhancing the overall swarm capability \cite{alqudsi2024synchronous, saffre2021design, alqudsi2024coordinated}.

Swarm robotics has emerged as a promising approach for creating captivating drone light shows (DLShows) through the coordinated movements of multiple autonomous drones, creating visually stunning displays previously unimaginable. DLShows have distinct advantages over fireworks and other traditional display methods: drones are reusable, eco-friendly, cause no air or noise pollution, and offer indoor flexibility \cite{nar2022optimal}. Although the upfront cost of custom drone swarms is significant, they prove to be a sustainable, cost-effective investment that pays off after a few shows, making them an attractive choice for frequent light shows events \cite{nonami2016drone}.

However, achieving optimal allocation and trajectories for a swarm of drones in light shows formations poses several challenges. The primary challenge lies in determining the optimal assignment of tasks to individual drones within the swarm and generating collision-free trajectories for each member \cite{ashraf2020online}. The complexity of allocation and trajectory planning increases with the number of drones involved \cite{hu2020uav}, and generating collision-free, time parameterized trajectories for each individual drone as dynamic flying robots adds more challenges \cite{alqudsi2025integrated}.

The utilization of a team of drones and the allocation of tasks to achieve predefined goal destinations have been extensively studied in the field of multi-agent systems \cite{turpin2014capt}. Their work addresses the assignment and the straight-line trajectory planning problem for a team of unlabeled robots. Two algorithms are proposed: C-CAPT, a centralized approach ensuring collision-free optimal solutions, and D-CAPT a decentralized alternative providing suboptimal results. In the existing literature, multiple approaches have been explored to address the challenge of formation control. For instance, \cite{toksoz2019decentralized} introduces a decentralized swarm control mechanism for a team of quadrotors, providing formation, rotation, tracking, and collision avoidance capabilities. While their approach showcases significant achievements in control precision and coordination, the control rules are developed based on the feedback linearization technique, which may exhibit limitations when faced with real-world uncertainties, such as wind disturbances or sensor noise \cite{alqudsi2024optimal, bianchi2023robust}. Further research is required to address these challenges and enhance system resilience in dynamic environments \cite{alqudsi2024analysis}. Nonetheless, their stability analysis and successful implementation provide valuable contributions to coordinating the complex movements required in DLShows.

Regarding motion coordination within drone swarms, \cite{nguyen2022quadrotor} introduces a formation tracking control method using fast terminal sliding mode control theory and a disturbance observer, showcasing improved resilience in the presence of disturbances. Contributing further to this field, \cite{pan2023distributed} outlines a distributed control scheme that leverages relative bearing and distance measurements to achieve unambiguous formation tracking. Another path planning method for swarm drone formations using an adapted artificial potential field approach is introduced in \cite{sun2020path}. The algorithm dynamically adjusts parameters to meet drone constraints, effectively addresses path oscillation, and proposes a target exchange technique for local optima. According to their simulation results, the suggested solution significantly advances multi-drone path planning in three dimensions. A recent study by \cite{nar2022optimal} introduced the Constrained Hungarian Method for Swarm Drones Assignment (CHungSDA), a method designed to optimally allocate multiple UAVs to waypoints. Building upon the Hungarian Algorithm (HA), they incorporated constraints tailored to drone shows applications. Although their simulation results exhibited promise, their focus was primarily on solving the optimal assignment problem without addressing the trajectory generation problem for providing dynamically feasible trajectories.

Expanding beyond formation control, \cite{fu2020formation} delves into obstacle avoidance within formation flying, combining position and speed consistency control with artificial potential field methods. Their approach generates, maintains, and reconstructs desired formations in both obstacle-free and obstacle-filled environments. In the field of DLShows, \cite{weng2022multi} presents a technique for generating multiple visual presentations using a visual hull approach, optimizing the process by minimizing the number of drones required for multi-view presentations and developing a flight algorithm for smooth animation and collision-free paths.

The aforementioned studies significantly contribute to the field of swarm control algorithms and formation tracking methods for drone applications, addressing important challenges such as coordination, collision avoidance, disturbance handling, and formation maintenance. However, most of these studies focus on individual aspects and do not simultaneously consider the assignment and trajectory generation problems, as well as the dynamics and optimal generation of time-parameterized trajectories for drones \cite{mcdonald2019real, alqudsi2021trajectory}. In light of these limitations, our research aims to fill this gap by proposing a Unified Assignment and Trajectory Generation (UATG) framework for a swarm of drones

The UATG framework introduces a unified approach that simultaneously addresses task allocation and trajectory generation, overcoming the limitations of the approaches prevalent in existing literature. This integrated methodology reduces computational overhead while ensuring the generation of trajectories that are computationally-efficient, collision-free, dynamically-feasible, and optimized for visual quality. Unlike existing methods, the UATG framework prioritizes both safety and robustness, which are critical for the success of DLShows. The framework's contribution is further demonstrated through its exceptional computational efficiency, enabling real-time coordination of large-scale "1000+" swarm drones on a standard laptop, representing a significant advancement over methods that struggle with scalability. By leveraging drone interchangeability, the framework enhances flexibility and robustness in dynamic environments, while maintaining inherent safety through robust collision avoidance and smooth, synchronized movements.

This paper conducts extensive simulations and experimental evaluations across diverse scenarios, validating the effectiveness and practical applicability of the proposed UATG framework. The results demonstrate the framework's ability to generate visually captivating and safe light show formations, ranging from basic patterns to complex 3D shapes involving thousands of drones, thereby highlighting the potential of swarm robotics for real-world DLShows. Furthermore, this research contributes to the advancement of multi-robot systems by addressing the UATG problem and providing insights into the challenges associated with optimal task allocation and 3D dynamic trajectory planning.

\section{System Architecture and Operational Framework}
In the domain of contemporary entertainment, the integration of technology and artistic expression has given rise to captivating DLShows. This section concisely outlines the essential components that constitute the sophisticated coordination of these displays, offering insights into their harmonious execution. Furthermore, it highlights the reasons behind the preference for quadrotors as the optimal platform chosen for these complex  aerial presentations.

\subsection{The Components of DLShows System}
\begin{figure}[ht]%
\centering
\includegraphics[width=0.5\textwidth]{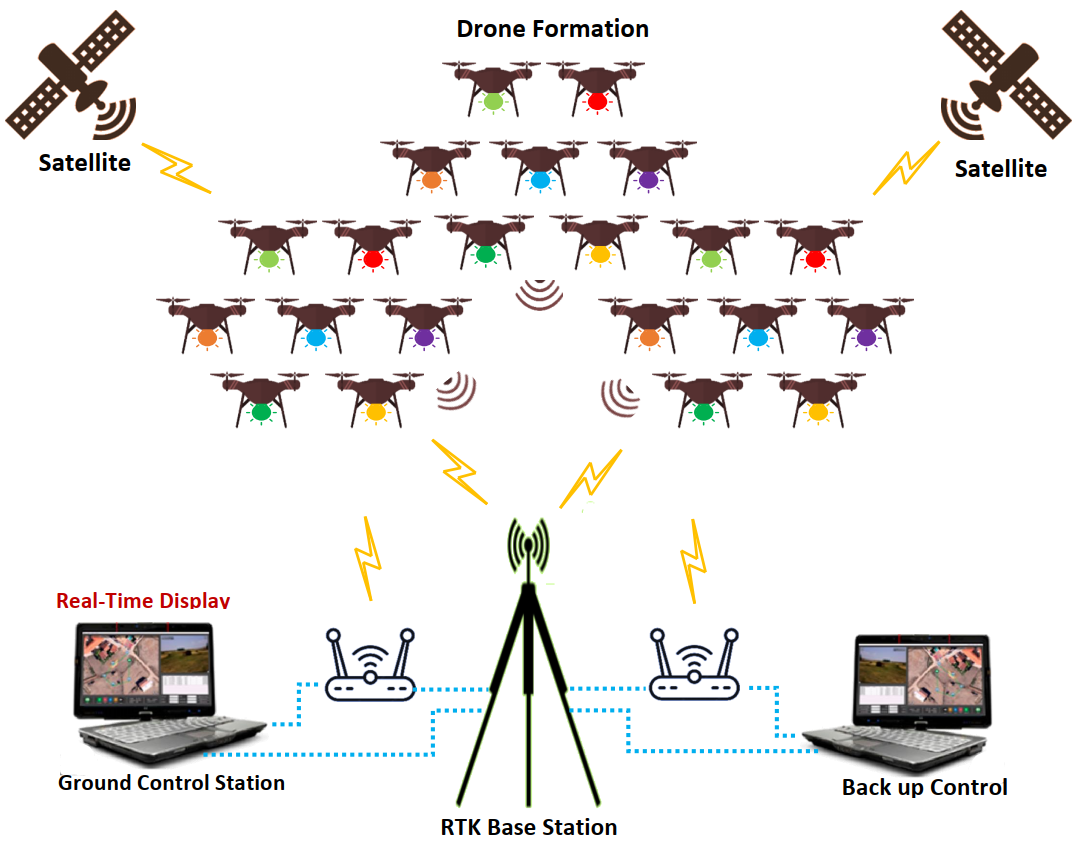}
\caption{Drone light show system architecture illustrating the integrated components: centralized control software, drone fleet with positioning and illumination systems, communication infrastructure, GCS, and redundancy mechanisms for reliable operation.}\label{droneShowComp}
\end{figure}

Aerial robots applications typically encompass two fundamental components: hardware and software. The hardware encompasses mechanical and electronic elements for flight, along with sensors for data acquisition and obstacle detection. Meanwhile, the software enhances flight stability, facilitates informed decision-making, enables obstacle avoidance, and ensures accurate navigation \cite{alqudsi2024advanced}. The DLShows system comprises several essential components that collaborate smoothly to create captivating aerial displays as depicted in Figure \ref{droneShowComp}. At the heart of this system is the Software $\&$ Control system, a centralized hub responsible for coordinating  the entire fleet of drones. This control system facilitates precise coordination, synchronization of movements, and the execution of complex choreographies that define the show's artistic appeal. The core performers of the DLShows are the drone fleet with equipment \cite{nar2022optimal}. These drones are equipped with essential elements, including GPS for accurate positioning, Light Emitting Diode (LED) lights for creating magical visual patterns, and robust power supply systems to ensure extended flight durations. The Communication System serves as the lifeline between the control center and individual drones, enabling real-time data exchange and enabling synchronized flight maneuvers. A dedicated Ground Control Station (GCS) provides operators with a central location for show management, data analysis, and monitoring of each drone's status. Additionally, the system incorporates a backup and redundancy mechanism to ensure reliability by having spare drones and redundant systems ready to smoothly take over in case of technical issues, ensuring uninterrupted performances.

\subsection{Quadrotors as Show Platforms}
Quadrotors are the preferred platform for DLShows due to their agility, precision, and stability. Their four-rotor design allows precise control over multi-directional movements, enabling complex aerial choreography essential for captivating displays \cite{alqudsi2021robust}. Quadrotors excel in maintaining formation even in challenging conditions, such as wind, ensuring consistency in performances. Their versatility accommodates a wide range of flight patterns, offering creative adaptability to different artistic visions. They are also known for their safety, suitable for indoor and outdoor shows, and customizable with specialized lighting and control systems for diverse visual effects \cite{nar2022optimal}. They scale effortlessly for various production sizes, simplifying choreography, and have a proven track record in high-profile events, establishing them as the ideal platform for innovative entertainment. For a comprehensive understanding of quadrotor dynamics, including their suitability for aerial performances, refer to the detailed insights provided in \cite{alqudsi2021robust}.

\subsection{Drone Light Shows Performance Workflow}
\begin{figure}[ht]%
\centering
\includegraphics[width=0.5\textwidth]{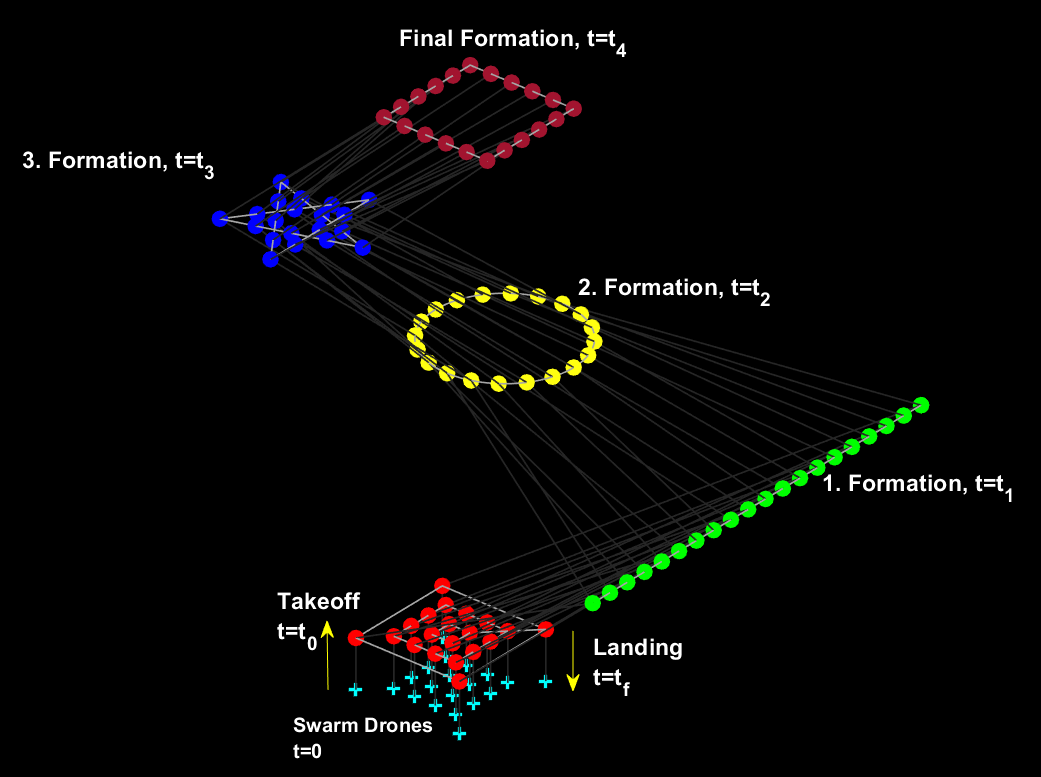}
\caption{Workflow of a DLShow performance illustrating the sequential phases. The sequence begins with synchronized takeoff and formation initialization, followed by smooth transitions between formations at $t=\{t_1, t_2, t_3, t_4\}$ with LED illumination patterns. Pre-programmed trajectory guidance ensures formation dynamics, concluding with a controlled landing.}\label{droneshow_seq}
\end{figure}

This section provides an in-depth analysis of the systematic operational sequence governing the execution of a DLShows performance. The complex coordination of drone movements within the aerial domain is systematically illustrated, encompassing the entire trajectory from takeoff initiation to the landing phase as depicted in Figure \ref{droneshow_seq}.

\begin{itemize}
    \item Takeoff and Formation Initialization: 
    The operational phase begins with the synchronized takeoff of individual quadrotors. Their ascension into the aerial domain is coupled with the establishment of a coherent formation, which sets the fundamental framework for the ensuing flight pattern.

    \item Pre-programmed Trajectory Guidance:
    The flight plan, carefully pre-calibrated, serves as the principal navigational directive for the drones. This algorithmically designed trajectory outlines the precise aerial path, encompassing maneuver profiles, dynamic transformations, and transitional sequences.

    \item Formation Dynamics:
    The drones smoothly execute coordinated maneuvers, yielding structured formations and configurations in the airspace. This phase encompasses the precise execution of geometric arrangements and complex  trajectories, contributing to the systematic evolution of aerial displays.

    \item LED Illumination Dynamics:
    The drones' visual impact is augmented through the complex modulation of their onboard LED arrays. These elements are coordinated in a harmonized manner, influencing luminosity patterns and spatial illumination, thereby augmenting the visual experience \cite{weng2022multi}.

    \item Transitions and Transformations:
    This phase is marked by smooth transitions between various formations, facilitated by the dynamic movement of drones. This coordinated shift in drone positions results in intricate spatial arrangements, creating a fluid transformation of the aerial visual scene.

    \item Final Formation and Landing:
    The show concludes with a striking formation for visual impact, followed by a controlled descent and precise landing of the drones.
\end{itemize}

\section{Theoretical Foundations of Swarm Task Assignment}
The coordination of drone swarms for light shows is fundamentally governed by task assignment principles from combinatorial optimization (CO), which involves optimally allocating a set of waypoints to a set of drones. This problem represents a cornerstone of CO with applications spanning multi-agent systems \cite{burkard1999linear}. In DLShows, this translates to assigning each drone to specific positions within complex aerial formations.

A critical property that fundamentally simplifies this assignment problem in homogeneous swarms is interchangeability. In multi-robot systems, agents are considered interchangeable when any agent can successfully fulfill the objectives of any given task without affecting mission outcomes \cite{sarcinelli2023control}. This represents a defining characteristic of DLShows, where fleets of identical UAVs deploy to form sophisticated, synchronized formations. The concept introduces valuable symmetry into the assignment problem, providing significant degrees of freedom that enable optimization for global performance metrics like total transition distance or energy consumption, rather than being constrained by individual drone capabilities.

The core assignment challenge in DLShows can be effectively formulated as a Linear Assignment Problem (LAP), seeking a one-to-one assignment between N drones and M waypoints that minimizes total cost, typically defined as the sum of individual assignment costs or distances traveled \cite{burkard1999linear}. Multiple algorithms optimally solve LAP, with the HA demonstrating particular suitability for large-scale drone swarms due to its polynomial-time complexity of $O(\max(N,M)^3)$ \cite{ismail2017decentralized}. This scalability establishes HA as a superior choice over alternatives including the Auction Algorithm (AA), whose computation time grows prohibitively for large N \cite{shi2020auction, li2022fast, ismail2017decentralized}, and metaheuristics like Genetic Algorithms (GAs), Ant Colony Optimization (ACO), and Particle Swarm Optimization (PSO), which often require extended computation times to reach near-optimal solutions \cite{alqudsi2025injected, lorena2002constructive, belkadi2015particle}. The Branch and Bound (B\&B) algorithm is another approach that considered as one of the well-known approaches for finding optimal solutions in CO problems \cite{morrison2016branch}. Despite being a powerful approach for tackling CO problems, it exhibits higher computational complexity compared to the HA, particularly when dealing with large-scale problem instances. The effectiveness of the HA has further inspired specialized adaptations like the CHungSDA, underscoring its relevance to aerial display applications. 
Consequently, the HA provides an efficient and optimal foundation for the task allocation component within the proposed UATG framework. Its ability to rapidly solve LAP for hundreds of interchangeable drones, finding not just feasible but optimally efficient assignments that ensure collision-free and visually smooth transformations between formations, represents a critical enabler for real-time generation of complex, large-scale aerial displays.

\section{Unified Assignment and Trajectory Generation Framework}
Assigning tasks to a team of drones and generating collision-free, time-parameterized trajectories adds even more challenges to be considered when dealing with a team of dynamic robots. This section introduces the UATG approach for drone swarms, addressing difficulties in swarm movement problems. UATG aims to ensure safety and visual quality in swarm formation performances by simultaneously allocating tasks and generating collision-free trajectories. When the number of drones (N) is greater than goals (M), the algorithm selects drones such that minimizing the overall mission completion cost. While the UATG framework will be designed to be applicable for any combination of the number of drones and goals, it's worth mentioning that this study, focusing on DLShows, assumes implicitly that the number of drones ($N$) is greater than or equal to the number of goals ($M$).

To achieve an optimal solution for the UATG problem involving a team of drones, careful selection of task allocations is essential as it directly impacts planning performance and overall efficiency. Among the various categories of task allocation problems, the LAP is widely recognized and will be implemented in this study to address the UATG problem \cite{burkard1999linear}. In the context of this paper, the LAP can be defined as follows: given N tasks to be performed by N drones, where each drone has a unique cost associated with completing its assigned task, the objective is to find the allocation that minimizes the total cost while ensuring that each goal is assigned to exactly one drone. Moreover, the time-parameterized trajectories derived from the LAP must satisfy two crucial criteria: they need to be both collision-free and dynamically feasible trajectories.

\subsection{Problem Definition, Modeling and Analysis}
\begin{figure}[ht]%
\centering
\includegraphics[width=0.35\textwidth]{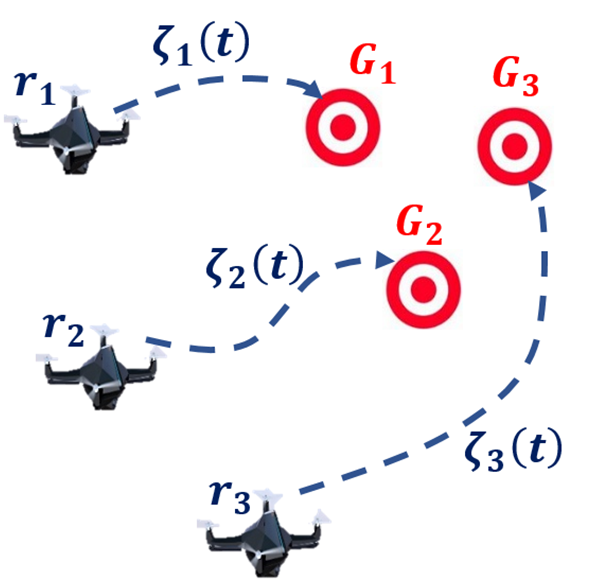}
\caption{UATG Problem Visualization showing initial drone positions ($r_i$), target goal locations ($G_i$), and generated 3D trajectories. The illustration highlights collision-free path planning and optimal assignment in obstacle-free environments.} \label{UATG_problem_settings}
\end{figure}

By incorporating the notion of interchangeability, the objective in the UATG problem is to simultaneously assign each goal by one drone and generate 3D trajectories to get the drones safely to their specified goals in the 3D environment as shown in Figure \ref{UATG_problem_settings}.

\subsubsection{Problem Definition}
The UATG problem can be defined as finding the optimal trajectories, denoted as $\zeta^* (t)$, that minimize a given cost functional. The cost functional is represented by the integral of a cost function, $L (\zeta (t), t)$, over the time interval from $t=t_0$ to $t=t_f$, as shown in Equation \ref{eq:eq1}:

\begin{equation}
\label{eq:eq1}
\zeta^*(t) = \substack{\text{arg min} \\ \zeta(t)} \int_{t=t_0}^{t=t_f} L(\zeta(t), t) \, dt
\end{equation}

Subjected to:
\begin{itemize}
    \item Valid Assignment: The assignment of tasks to drones must be valid, ensuring that each goal is assigned to exactly one drone. This ensures that all tasks are properly allocated among the drone team \cite{schwarzrock2018solving}.
    \item  Interchangeability: All drones are implicitly assumed to be interchangeable and homogeneous, with no preference for specific goal locations or prioritization of certain drones over others.
    \item Terminal Conditions: The trajectories of the drones must satisfy the given initial and final conditions, ensuring that the drones begin and finish their assigned tasks correctly \cite{alqudsi2023general}.
    \item Collision Avoidance: The trajectories of the drones should be planned in a way that avoids collisions between the drones and any other obstacles (if any) in the flying environment. This is crucial for the smooth execution of tasks and preventing any potential damage or interference. It's worth noting that for DLShows, the implicit assumption is that the flying environment remains free of obstacles.
    
    \item Drone Capabilities: Trajectories must account for individual drone capabilities and constraints, such as maximum  speed, actuators' limits, acceleration, and maneuverability. Incorporating these factors ensures that the trajectories are feasible for execution by the drones \cite{hanover2024autonomous}.
\end{itemize}

\BlankLine
\subsubsection{Problem Decomposition}
\label{subsubsec:problem-decomposition}

The UATG framework decomposes the overall coordination challenge into two fundamental subproblems that require joint optimization:

\paragraph{Optimal Task Assignment}
The algorithm computes an optimal assignment of $N$ drones to $M$ goal positions by constructing a binary assignment matrix $\Phi \in \{0,1\}^{N \times M}$; once "if drone $i$ is assigned to goal" and zeros "otherwise".
The assignment satisfies the constraint that each goal is assigned to exactly one drone when $N \geq M$.

To effectively employ the HA in solving the LAP within the UATG problem, it becomes crucial to establish a cost matrix that the algorithm can utilize for obtaining the optimal assignment. The creation of the cost matrix depends on the chosen optimization criteria, such as minimizing the sum of distances between each drone and all goals or minimizing the sum of distances-squared, ... etc. The selection of the tasks assignment optimization-criteria directly impacts the overall performance of the UATG problem. Using sum of distances criterion may not guarantee Drone-to-Drone ($D2D$) collision-free while the drones travelling to their destinations \cite{turpin2014capt}. Within the scop of this study, the optimization-criteria will be chosen as follows: finding the optimal assignment that minimizes the sum of the squared distances traveled by all drones to their respective assigned destinations, which corresponds to selecting paths result in the minimum velocity norm squared throughout the entire flight duration  \cite{naidu2002optimal}. Therefore, the cost matrix will be determined according to Equation \ref{eq:eq4}.

\begin{equation}
\label{eq:eq4}
\boldsymbol{B}^*(t) = \substack{\text{arg min} \\ \zeta(t)} \int_{t=t_0}^{t=t_f} |\dot{\zeta}(t)|^2 \, dt
\end{equation}

Where, $\dot{\zeta}(t)$ represents the velocity components. The Euclidean distance metric provides an efficient heuristic for initial assignment optimization in obstacle-free environments typical of DLShows. These initial position-to-goal connections serve as waypoints for constructing dynamically feasible trajectories through the algorithm introduced in \cite{alqudsi2023numerically}, which generates smooth, optimized paths respecting quadrotor dynamics and constraints. Consequently, Equation \ref{eq:eq4} can be reformulated as follows:
\begin{equation}
\label{eq:eq5}
B_{ij} = \|X_i^R - X_j^G\|^2, \quad i=1:N, \quad j=1:M
\end{equation}

Where \(X_i^R\) and \(X_j^G\) represent the initial position of drone \(i\) and the location of goal \(j\), respectively. \(B\) is a matrix in \(\mathbb{R}^{N \times M}\), and \(B_{ij}\) denotes the element of the cost matrix corresponding to row \(i\) and column \(j\).

\BlankLine
\paragraph{Dynamically Feasible Trajectory Planning}
For each drone-goal assignment, the framework generates time-parameterized trajectories that satisfy three critical requirements:

\begin{enumerate}
    \item Trajectory Optimality: The generated paths must minimize the cost functional defined in Equation \ref{eq:eq1}. The trajectory generation employs $k^{th}$-order polynomial segments between waypoints:
    \begin{equation}
    \zeta(t) = \sum_{i=0}^{k} c_i t^i
    \end{equation}
    where coefficients $c_i$ are optimized for minimum snap criteria to ensure smooth visual transitions:
    \begin{equation}
    J = \int_{t_0}^{t_f} \left(\frac{d^4 \zeta}{dt^4}\right)^2 dt
    \end{equation}
    This formulation guarantees $C^3$ continuity while respecting dynamic constraints:
    \begin{equation}
    |\dot{\zeta}(t)| \leq v_{\text{max}}, \quad
    |\ddot{\zeta}(t)| \leq a_{\text{max}}, \quad
    |\dddot{\zeta}(t)| \leq j_{\text{max}}
    \end{equation}
    
    \item Collision Avoidance: In obstacle-free environments characteristic of DLShows, the primary safety consideration is inter-drone collision avoidance. The constraint ensures minimum separation between all drone pairs:
    \begin{equation}
    \label{eq:eq3}
    \| \chi_i(t) - \chi_j(t) \| \geq 2R\eta, \quad \eta > 1, \quad \forall t \in [t_0, t_f]
    \end{equation}
    where $\| \chi_i(t) - \chi_j(t) \|$ denotes the Euclidean distance between drones $i$ and $j$, $R$ represents the effective radius of a quadrotor (typically the larger moment arm), and $\eta > 1$ provides a safety margin accounting for aerodynamic interference and localization uncertainties.
    
    \item Dynamic Feasibility: Generated trajectories must respect the inherent dynamic constraints of quadrotor systems. Unlike simplified kinematic models, quadrotors cannot instantaneously change velocity or direction, necessitating smooth, continuously differentiable paths that account for actuator limits and vehicle dynamics \cite{alqudsi2023general, klarin2020drone}.
\end{enumerate}

\subsection{Trajectory Generation Process}
The trajectory generation module of UATG extends a customized version of the trajectory generation and optimization (TGO) framework \cite{alqudsi2023numerically} to produce time-parameterized, dynamically feasible trajectories for each agent in a swarm. 
For drone \(i\) the planned trajectory is represented as a piecewise polynomial
\[
\mathbf{p}_i(t) = 
\begin{cases}
\displaystyle \sum_{m=0}^{k} \mathbf{c}_{i,1,m}\,(t-t_0)^{m}, & t\in[t_0,t_1],\\[6pt]
\displaystyle \sum_{m=0}^{k} \mathbf{c}_{i,2,m}\,(t-t_1)^{m}, & t\in[t_1,t_2],\\
\;\vdots &
\end{cases}
\]
where \(\mathbf{c}_{i,s,m}\in\mathbb{R}^3\) are the vector coefficients for segment \(s\) and polynomial order \(k\). Coefficients are chosen so that \(\mathbf{p}_i(t)\) is smooth and differentiable up to the required order (we enforce continuity of position, velocity, acceleration and jerk at segment boundaries to ensure smooth snap behavior). Since quadrotor drones are the chosen show platform, the polynomials are designed to be continuous and differentiable up to the fourth derivative (snap), allowing the drones to follow the generated trajectories without abrupt changes in velocity or acceleration. This is critical for maintaining formation stability and achieving the high visual quality required in DLShows.

The trajectory synthesis is posed as a constrained optimization problem that minimizes a quadratic cost expressing trajectory smoothness and energetic economy. A commonly used formulation in our work is the minimum-snap cost
\begin{equation}
\min_{\{\mathbf{c}_{i,s,m}\}} \; J \;=\; \sum_{i=1}^{N}\sum_{s}\int_{t_{s}}^{t_{s+1}} \big\| \tfrac{\mathrm{d}^4}{\mathrm{d}t^4}\mathbf{p}_i(t)\big\|^2 \,\mathrm{d}t,
\label{eq:minsnap}
\end{equation}
which is quadratic in the polynomial coefficients. In matrix form this leads to the canonical quadratic program
\[
\min_{\mathbf{x}} \; \tfrac{1}{2}\mathbf{x}^\top H \mathbf{x} + \mathbf{f}^\top\mathbf{x}
\quad\text{s.t.}\quad A\mathbf{x}=\mathbf{b},\; G\mathbf{x}\le \mathbf{h},
\]
where \(\mathbf{x}\) stacks all polynomial coefficients for every drone, \(H\) is positive semidefinite (assembled from integrals of basis-function derivatives), and \(A,\mathbf{b}\) encode equality constraints.
\BlankLine
The optimization is subject to the following practical and safety constraints:
\begin{itemize}
  \item Waypoint constraints: \(\mathbf{p}_i(t_{s}^\pm)=\mathbf{w}_{i,s}\) for prescribed waypoint locations \(\mathbf{w}_{i,s}\) and times \(t_s\).
  \item Continuity constraints: continuity of \(\mathbf{p}_i,\dot{\mathbf{p}}_i,\ddot{\mathbf{p}}_i,\dddot{\mathbf{p}}_i\) at segment boundaries.
  \item Dynamic limits: \(\|\dot{\mathbf{p}}_i(t)\|\le v_{\max},\; \|\ddot{\mathbf{p}}_i(t)\|\le a_{\max}\), and actuator-dependent bounds (expressed as linear or convex constraints on coefficients).
  \item Collision-avoidance: inter-drone separation \(\|\mathbf{p}_i(t)-\mathbf{p}_j(t)\|\ge d_{\min}\) for all \(i\neq j\) and \(t\in[t_0,t_f]\) (Equation~\ref{eq:eq3}). In practice, this nonconvex constraint is handled by conservative convexification: (i) enforcing separation at a discrete set of time samples, (ii) using safe corridors or separating hyperplanes, or (iii) applying iterative linearization and re-planning when predicted violations occur.
  \item Performance timing and synchronization: trajectory end-times and waypoint timings are constrained to meet show timing \(t_{\mathrm{stage}}\) and to synchronize LED illumination sequences, i.e. \(\mathbf{p}_i(t_{\mathrm{sync}})\) must coincide with illumination events.
\end{itemize}

Because the cost in \eqref{eq:minsnap} is quadratic in the coefficients and the waypoint/continuity constraints are linear, the discretized problem is a convex quadratic program (QP) when collision conditions are enforced via linearized or discretized constraints. This structure allows reliable global optimization via standard QP solvers. For numerical robustness and conditioning we represent segments in a well-conditioned polynomial basis (e.g. shifted Legendre, Bernstein or B-spline basis) and employ scaling of time intervals; these measures reduce numerical ill-conditioning when \(k\) or the number of waypoints \(W\) is large \cite{alqudsi2023numerically}.

Leveraging this convex optimization foundation, the UATG framework implements a comprehensive trajectory generation pipeline that integrates several practical components: (i) automatic conversion of goal-point clouds into waypoint sequences, (ii) explicit enforcement of actuator and inertia limits so trajectories are dynamically feasible, (iii) a re-planning loop that locally perturbs waypoint timings and reconducts the QP when potential D2D conflicts are detected, and (iv) timing constraints that align LED modulation with drone motion. Together, these components produce smooth, energy-efficient trajectories that satisfy safety, timing and dynamic requirements for large-scale DLShows.

\subsection{Problem Solving Procedures}
To solve both the assignment and planning problems concurrently, the UATG framework, outlined in Algorithm~(\ref{UATG_algo}), follows a straightforward procedure. The algorithm begins by defining the number and initial configurations of swarm drones, along with timing and formation characteristics. Moving forward, the assignment problem is addressed. The algorithm then constructs the cost matrix $\boldsymbol{B}$ based on selected optimization criteria, subsequently employing it in the next stage of the algorithm to optimally solve the LAP.
\begin{algorithm}
\SetAlgoLined
\KwIn{Drone set $\mathcal{D} = \{1,\dots,N\}$, goal set $\mathcal{G} = \{1,\dots,M\}$, initial states $\mathbf{X}^R$, goal positions $\mathbf{X}^G$, safety parameters $(R, \eta)$, dynamic constraints}
\KwOut{Assignment matrix $\Phi$, Optimal trajectory set $\{\mathbf{p}_i(t)\}_{i=1}^N$}
\caption{A Pseudocode of the UATG Framework}\label{UATG_algo}
    \BlankLine
    \textbf{Initialization:}\;
    Load choreography $\mathcal{C}$ and timing parameters $t_{\text{stage}}, t_{\text{sync}}$\;
    Compute formation states $\mathcal{S}_{\text{stage}}$ for all stages\;
    \BlankLine
    \textbf{Assignment:}\;
    \Indp
    Construct cost matrix $\mathbf{B} \in \mathbb{R}^{N \times M}$ where $B_{ij} = \|\mathbf{X}_i^R - \mathbf{X}_j^G\|^2$\;
    Solve LAP: $\Phi^* = \underset{\Phi}{\arg\min} \sum_{i=1}^N \sum_{j=1}^M \phi_{ij} B_{ij}$\;
    Subject to: $\sum_{i=1}^N \phi_{ij} = 1$ $\forall j$, $\sum_{j=1}^M \phi_{ij} = 1$ $\forall i$, $\phi_{ij} \in \{0,1\}$\;
    Extract optimal assignment mapping: $\mathcal{A}: \{1,\dots,N\} \rightarrow \{1,\dots,M\}$\;
    \Indm
    \BlankLine
    \textbf{Trajectory Generation:}\;
    \Indp
    \ForEach{drone $i \in \{1,\dots,N\}$}{
        Determine waypoint sequence $\mathcal{W}_i = \{\mathbf{w}_{i,1}, \mathbf{w}_{i,2}, \dots, \mathbf{w}_{i,W}\}$\;
        Formulate minimum-derivative optimization:
        \[
        \min_{\mathbf{c}_{i,s,m}} \sum_{s=1}^{W-1} \int_{t_s}^{t_{s+1}} \left\| \frac{d^4\mathbf{p}_i(t)}{dt^4} \right\|^2 dt
        \]
        Subject to:
        \begin{itemize}
            \item Waypoint constraints: $\mathbf{p}_i(t_s) = \mathbf{w}_{i,s}$
            \item Boundary conditions: $\mathbf{p}_i(t_0) = \mathbf{X}_i^R$, $\mathbf{p}_i(t_f) = \mathbf{X}_{\mathcal{A}(i)}^G$
            \item Dynamic constraints: $\|\dot{\mathbf{p}}_i(t)\| \leq v_{\max}$, $\|\ddot{\mathbf{p}}_i(t)\| \leq a_{\max}$, $\|\dddot{\mathbf{p}}_i(t)\| \leq j_{\max}$
            \item Continuity: $\mathbf{p}_i$, $\dot{\mathbf{p}}_i$, $\ddot{\mathbf{p}}_i$, $\dddot{\mathbf{p}}_i$ continuous at waypoints
        \end{itemize}
        Solve QP to obtain trajectory coefficients $\{\mathbf{c}_{i,s,m}\}$\;
    }
    \Indm
    \BlankLine
    \textbf{Collision Verification:}\;
    \While{$\exists i,j: \min_t \|\mathbf{p}_i(t) - \mathbf{p}_j(t)\| < 2R\eta$}{
        Apply spatiotemporal perturbation to conflicting trajectories\;
        Re-optimize trajectories locally\;
    }
    \BlankLine
    \textbf{Execution and Synchronization:}\;
    \Indp
    - Integrate LED control signals with trajectory timing\;
    - Implement real-time monitoring and contingency handling\;
    - Execute synchronized swarm performance\;
    \Indm
    \BlankLine
    \Return $\Phi^*, \{\mathbf{p}_i(t)\}_{i=1}^N$, performance metrics\;
\end{algorithm}

After resolving the assignment phase, the algorithm proceeds to generate time-parametrized trajectories for each drone-to-goal pair. This is achieved by implementing a customized version of the TGO algorithm introduced in \cite{alqudsi2023numerically}. The TGO algorithm facilitates agile drone flight while considering constraints related to robot dynamics, actuator inputs, and the flight environment. It employs time-parametrized polynomial trajectories based on a predefined sequence of waypoints, ensuring dynamically feasible and collision-free trajectories.

In the next stage, the algorithm determines dynamic trajectories while considering the capabilities of the drones. To ensure collision-free movements during trajectory execution, the algorithm performs a double check for the possibility of D2D collisions that may arise during dynamic trajectory generation. This verification step guarantees the safe execution of designed trajectories, mitigating the risk of collisions among the drones. By following this systematic procedure, the proposed approach addresses both the assignment and planning problems in a cohesive manner, enabling the optimal coordination and path planning for the team of drones.

\subsection{Theoretical Analysis and Algorithm Properties}
The computational efficiency, convergence guarantees, and robustness of the proposed UATG algorithm are key aspects that underpin its practical applicability for large-scale DLShows. From a complexity standpoint, the assignment problem is solved with a computational complexity of $O(\max(N,M)^3)$. Once the trajectory-constrained assignment stage is completed, trajectory generation is performed with a complexity of $O(N \cdot W \cdot k^3)$, where $W$ denotes the number of waypoints and $k$ the order of the polynomial used in the trajectory representation. Additionally, collision checking between drones is carried out with a complexity of $O(N^2)$, which can be further improved by applying spatial partitioning techniques.

With respect to optimality, the HA guarantees globally optimal solutions for the linear assignment formulation. Similarly, the trajectory optimization process is formulated as a convex quadratic program, which ensures global optimality for the chosen smoothing criteria while strictly enforcing the dynamic and safety constraints. This dual-stage optimization provides both optimal allocation of drones to goals and dynamically feasible trajectories.

The UATG algorithm also incorporates important safety and robustness properties. Collision avoidance is guaranteed through the hard distance constraints between homogeneous drones defined in Equation~\ref{eq:eq3}, while dynamic feasibility is maintained by explicitly enforcing actuator and physical limitations during trajectory generation. Furthermore, the concept of interchangeability within the swarm provides robustness against individual drone failures, as the system can reallocate goals without compromising the overall formation performance. These properties collectively ensure that the UATG algorithm can reliably coordinate large drone swarms in complex, safety-critical aerial displays.

\section{Experimental Validation and Performance Analysis}
This section demonstrates the effectiveness of the proposed UATG framework in achieving optimal performance, ensuring inter-drone collision avoidance, and enhancing the overall flight dynamics of a drone swarm. To do so, we implemented and evaluated the proposed UATG framework in a simulated environment that closely emulates real-world conditions for DLShows. This includes considerations for gravitational effects and aerodynamics interactions between drones. While certain aspects like wind effects are not explicitly modeled in the simulation, they are implicitly addressed through the robust hybrid controller, as detailed in \cite{alqudsi2021robust}. The controller is designed to account for uncertainties, whether due to noise, external factors such as wind, or model uncertainties. We conducted several experiments, varying fleet sizes and designing formations in a 3D coordinate system environment to validate the robustness and effectiveness of our simulation in replicating realistic DLShows performances. The computational capabilities of the hardware as the primary limitation imposed on the simulation will be mention in the subsequent subsection. 

\subsection{Simulation Environment and Parameters}
To thoroughly assess the UATG framework, the simulations were conducted in a Software-in-the-Loop (SITL) environment. This setup used Python 3.10, MATLAB R2022b, and Gazebo software to replicate real-world conditions. The simulations were run on a standard laptop with the following specifications: an Intel Core i7-1065G7 processor running at 1.30GHz, an NVIDIA GeForce GTX 1650 GPU with 4 GB of dedicated memory, and 16 GB of RAM.

This research incorporated a realistic drone model, considering parameters like aerodynamic and drag coefficients, dimensions, maximum rotor speed, and flight time as outlined in Table \ref{tab:quadrotor-params}. These parameters were carefully chosen to accurately capture the dynamic behavior of the drones during swarm formations and transitions. The simulation environment, designed to replicate real-world conditions, featured a 3D coordinate system without obstacles, aligning with the UATG framework's settings. We systematically varied parameters such as fleet size and formation complexity, utilizing a range of metrics, including formation completion time, trajectory smoothness, inter-drone distances, control effort, and assignment optimality, to provide a thorough evaluation of the algorithm's capabilities and effectiveness.

\begin{table}[htbp]
    \centering
    \caption{Quadrotor Model Parameters for Experimental Simulations}
    \label{tab:quadrotor-params}
    \begin{tabular}{|l|l|}
    \hline
    Parameter & Value/Unit \\
    \hline
    Quadrotor total mass & 1.79 kg \\
    Motor to c.g. distance ($R$) & 0.18 m \\
    Roll and pitch inertia & $1.335 \times 10^{-2}$ kg$\cdot$m$^2$ \\
    Yaw inertia & $2.465 \times 10^{-2}$ kg$\cdot$m$^2$ \\
    Aerodynamic coefficient & $8.82 \times 10^{-6}$ N/(rad/sec)$^2$ \\
    Drag coefficient & $1.09 \times 10^{-7}$ N$\cdot$m/(rad/sec)$^2$ \\
    Max rotor speed & 860.1681 rad/sec \\
    Max flight time & 28 min \\
    Quadrotor configuration & Plus-Config. \\
    \hline
    \end{tabular}
\end{table}

\subsection{Results and Discussion}
This section delves into the results of our experiments, providing an in-depth discussion of the implications and strengths of the UATG framework in the context of swarm formations and DLShows performances. The algorithm's performance was evaluated across various scenarios, demonstrating its robustness and versatility.

The initial positions of the swarm drones can be set randomly or arranged in a square mesh on the ground. A safe distance of $1.5$ meters between drones, i.e. $\eta = 0.75/R$, is chosen to consider the effects of aerodynamic interference and real-world uncertainties. The hovering time is chosen to be $2$ seconds for each formation stage. To define the desired formations created by the swarm drones, we utilized our developed auxiliary software. This software converts 3D shapes, words, numbers, letters, or pictures into coordinates and waypoints in the 3D space, providing precise instructions for the UATG framework.

\begin{figure*}[htbp]%
\centering
\includegraphics[width=0.9\textwidth]{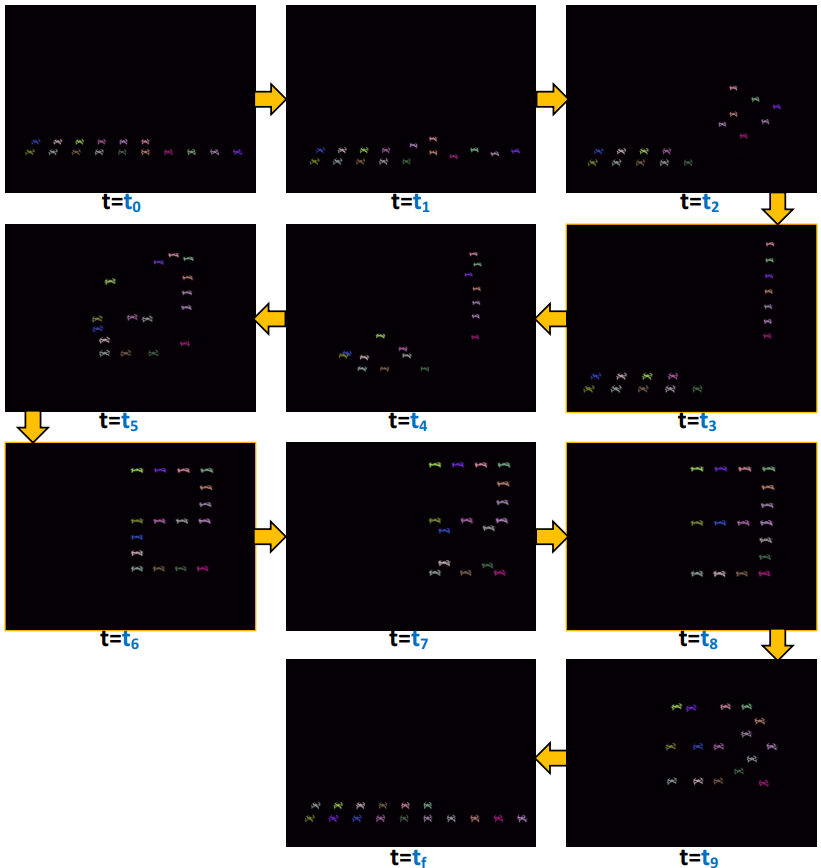}
\caption{Sequential numerical formations demonstrating the UATG algorithm's capability to transform a 16-drone swarm through digits 1, 2, and 3. The sequence illustrates smooth transitions between formations while maintaining optimal assignments and collision-free trajectories.} \label{sim_numbers}
\end{figure*}

\BlankLine
\paragraph{\textbf{Formation 1: Numerical Digit Sequences}}
In the first scenario, we used the UATG framework to choreograph the synchronized formation of numerical digits (1, 2, and 3) using a fleet of 16 drones. The primary aim was to showcase the algorithm's proficiency in efficiently assigning global optimal waypoints, enabling the drones to transition seamlessly between different numerical formations. This scenario simulated a DLShows where numerical themes are artistically displayed in the night sky, as depicted in Figure  \ref{sim_numbers}.

The simulation of this scenario consisted of multiple scenes, each representing a specific phase in the performance. Let's delve into these scenes to better understand the process:

    \begin{itemize}
        \item Takeoff ($t=t_0$):
        The simulation starts with the drones taking off from their initial positions. This phase marks the beginning of the performance.

        \item Formation 1 ($t=t_1$ to $t=t_3$):
        The drones then transition to the first formation, shaping the digit "1." Notably, only the drones that contribute to the optimal assignment for this formation actively participate, while the remaining drones maintain their hovering positions waiting for their turn. In the live performance, the drones not actively involved in a particular formation will deactivate their lights, creating a visually striking effect as they await their turn.

        \item Formation 2 ($t=t_4$ to $t=t_6$): Following the successful execution of the first formation, the drones move dynamically to create the shape of the digit "2." The UATG framework ensures that this transition is both quick and precise.

        \item Formation 3 ($t=t_7$ to $t=t_8$): Continuing the performance, the drones smoothly transform into the shape of the digit "3."

        \item Return to Hover ($t=t_9$): After captivating the audience with numerical formations, the drones conclude their performance by returning to hover collectively above the landing area. This phase is essential for a safe and coordinated landing.

        \item Autonomous Landing ($t=t_f$): Finally, the drones execute a controlled descent, ensuring they land with precision. This concludes the performance.

    \end{itemize}

The algorithm execution time of this scenario is 0.3885 seconds, emphasizing the UATG framework's efficiency in orchestrating visually captivating numerical formations while optimizing the drones' trajectories. This scenario provides a foundation for more complex and engaging DLShows with diverse themes and formations.

\begin{figure*}[htbp]%
\centering
\includegraphics[width=0.9\textwidth]{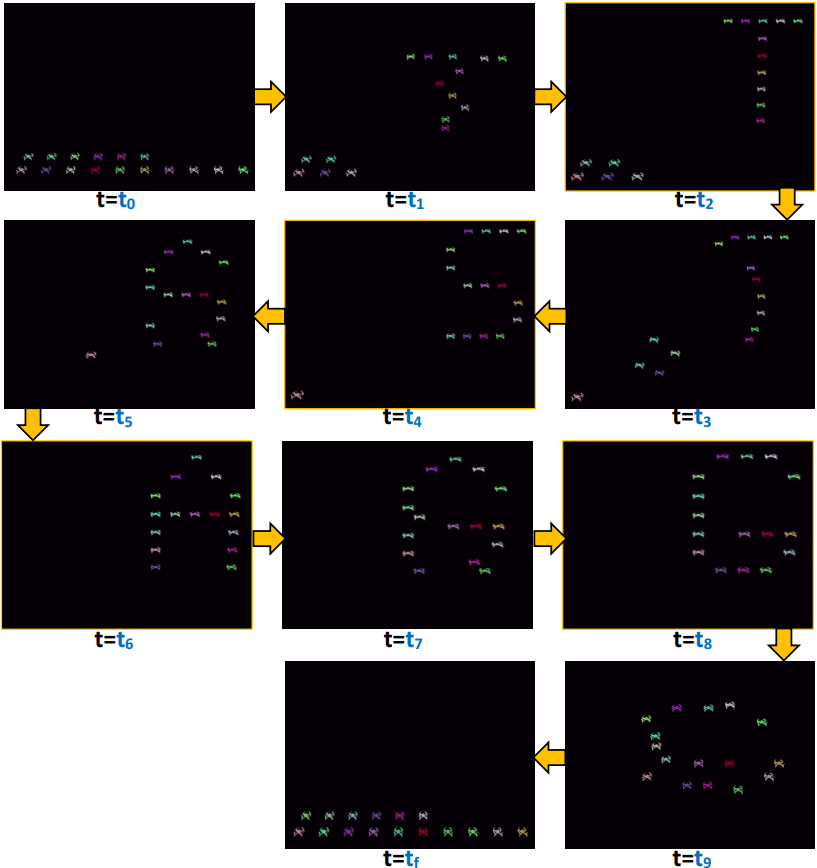}
\caption{Alphabetical character formations (T, S, A, G) generated by a 16-drone swarm, showcasing the algorithm's flexibility in creating complex letter-based patterns with efficient task allocation and synchronized movements.} \label{sim_letters}
\end{figure*}

\BlankLine
\paragraph{\textbf{Formation 2: Alphabetical Character Patterns}}
In this scenario, the drone swarm embarked on the artistic endeavor of crafting letters (T, S, A, and G) in a synchronized manner. This experimental simulation was conducted with a fleet of 16 drones, serving as a testament to the UATG framework's adaptability and creative potential in orchestrating complex letter-based formations. The proposed algorithm solved this scenario in 0.3440 seconds.
As depicted in Figure \ref{sim_letters}, the simulation vividly illustrates the algorithm's proficiency in orchestrating intricate formation designs, showcasing its remarkable flexibility in producing visually captivating letter-based arrangements.

To gain a deeper insight into the simulation's flow, let's explore the individual scenes that make up this scenario:
\begin{itemize}
    
    \item Takeoff ($t=t_0$): 
    The scenario kicks off with the drones taking flight from their initial positions, marking the commencement of this creative performance.

    \item Formation T ($t=t_1$ to$ t=t_2$): 
    The drones then transition into the formation of the letter "T." Notably, the UATG framework optimally selects the subset of drones required for this particular formation. Meanwhile, the other drones, which are not actively involved, would have their lights turned off in the real performance, symbolizing their readiness as they hover in position, patiently awaiting their turn to participate in the performance.

    \item Formation S ($t=t_3$ to $t=t_4$):
    Building on the previous formation, the drones elegantly transform into the shape of the letter "S." Once again, only the drones essential for this formation are actively engaged, aligning with the UATG framework's efficiency in selecting optimal drone assignments.

    \item Formation A ($t=t_5$ to $t=t_6$):
    The performance continues to unfold as the drones craft the letter "A" with precision.

    \item Formation G ($t=t_7$ to$ t=t_8$):
    The climax of the scenario arrives with the creation of the letter "G".
    
    \item Return to Hover ($t=t_9$):
    After the successful execution of captivating letter-based formations, the drones gracefully begin returning to hover collectively above the designated landing area. This phase is integral for ensuring a coordinated and safe landing.

    \item Autonomous Landing ($t=t_f$): 
    To conclude the performance, the drones execute a precisely controlled descent, ensuring a flawless and synchronized landing, thereby concluding this artistic drone display.
\end{itemize}

\BlankLine
\paragraph{\textbf{Formation 3: Geometric Shape Transformations}}
In this scenario, 43 drones worked together to craft formations of shapes, specifically a crescent moon and a star, in a synchronous manner.

\begin{figure*}[htbp]%
\centering
\includegraphics[width=0.8\textwidth]{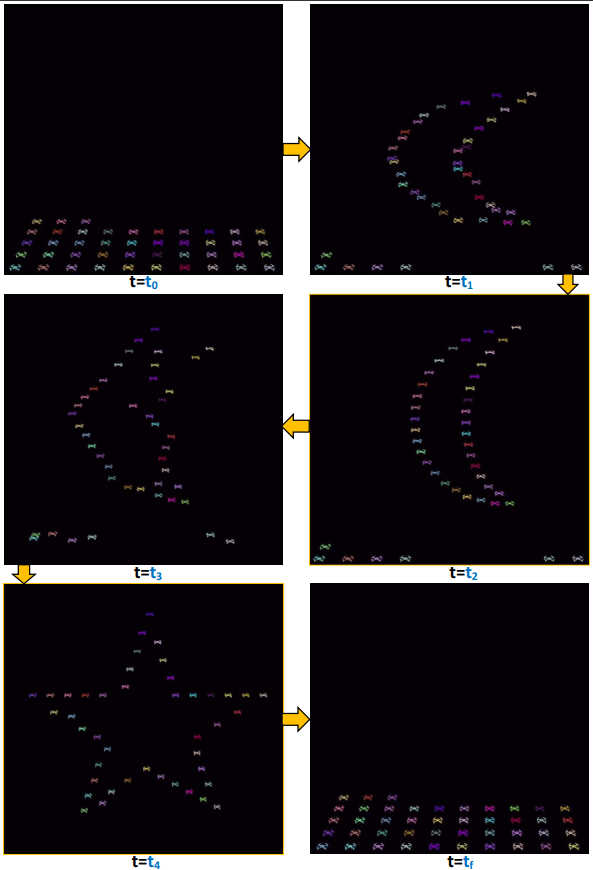}
\caption{Geometric shape transformations featuring a 43-drone swarm transitioning between crescent moon and star formations. The results demonstrate scalable coordination for medium-sized swarms executing precise 2D shape patterns.} \label{sim_shapes}
\end{figure*}

The process began with the drones taking off from their initial positions. Figure \ref{sim_shapes} illustrates six key scenes during this scenario, marked by time points $t=t_0, t=t_1, ..., t=t_4, and t=t_f$. The initial stage, $t=t_0$, involved the takeoff of the drones. At $t=t_1$, there was a transitional movement to form the shape of a crescent moon, which was successfully achieved at $t=t_2$. 
Subsequently, at $t=t_3$, the drones transitioned to form the shape of a star, which was accomplished at $t=t_4$. After successfully creating these formations, the drones then returned to hover above the designated landing area, marked by a transitional movement at $t=t_f$, and finally executed an autonomous landing.

This scenario, as depicted in figure \ref{sim_shapes}, exemplified the algorithm's ability to effectively coordinate a larger drone fleet to craft intricate and visually captivating shapes. The algorithm execution time of this scenario is 0.4913 seconds.

\begin{figure*}[htbp]%
\centering
\includegraphics[width=0.95\textwidth]{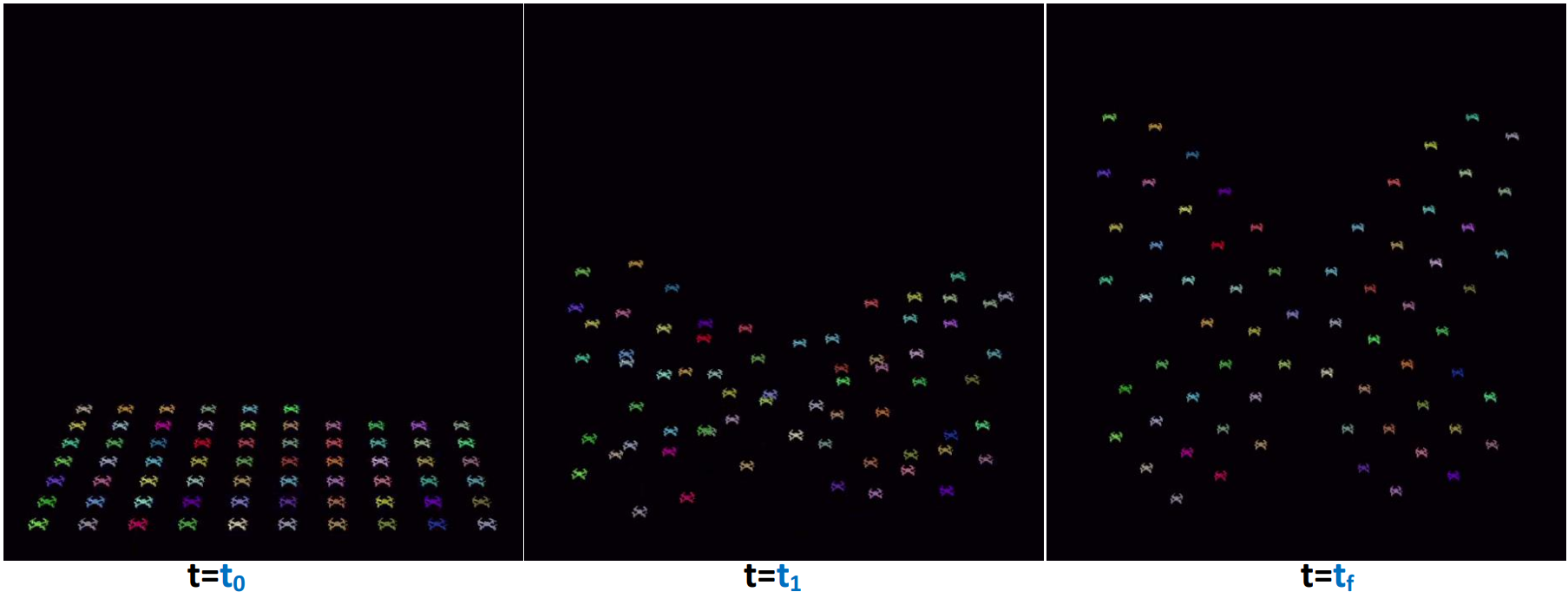}
\caption{Complex organic pattern formation showing a 66-drone swarm configured as a butterfly, highlighting the algorithm's capacity to orchestrate intricate biological shapes with smooth, visually appealing transitions.} \label{sim_Butterfly}
\end{figure*}

\BlankLine
\paragraph{\textbf{Formation 4: Complex Organic Pattern (Butterfly)}} 
In the fourth scenario, the spotlight was on crafting a visually intricate butterfly formation using a drone ensemble comprising 66 members. This experiment exemplified the UATG framework's prowess in orchestrating a substantial drone fleet to execute complex and captivating organic shapes. 
The simulation began with drones taking flight, forming a butterfly shape within the given timeframe. This showcases UATG's exceptional coordination of large drone ensembles for complex and visually mesmerizing displays. The butterfly formation in Figure \ref{sim_Butterfly} serves as a testament to UATG's ability to manage sizable drone fleets and choreograph elaborate light shows on a grand scale, emphasizing its potential for large-scale performances. The algorithm execution time of this scenario is 0.1373 seconds.

\begin{figure}[ht]%
\centering
\includegraphics[width=0.45\textwidth]{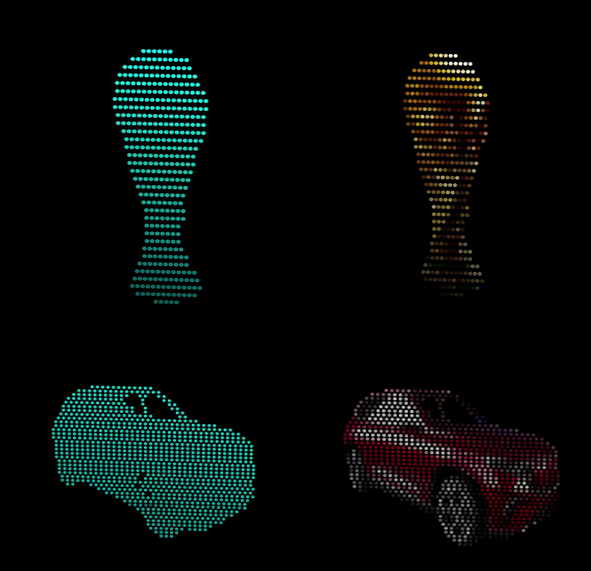}
\caption{Validation of the UATG framework on large-scale 3D structures. The upper panel shows a World Cup trophy formed by 389 drones, while the lower panel shows a TOGG car formed by 1008 drones, collectively demonstrating the algorithm's capability for high-density swarm coordination.} \label{worldcup_car}
\end{figure}

\BlankLine
\paragraph{\textbf{Formation 5: Large-Scale 3D Structures}} 
This Scenario aimed to form the World Cup and a 3D image for the TOGG Car using 389 and 1008 drones, respectively as shown in Figure \ref{worldcup_car}.
The drones gracefully ascended and formed intricate shapes representing the World Cup and the TOGG Car.  The algorithm execution time for the World Cup and the TOGG Car DLShows took 0.1896 and 1.5795 seconds, respectively. These formations were executed with precision and intricacy, exemplifying the UATG framework's remarkable capacity for orchestrating complex, highly detailed drone displays on a grand scale.
In another scenario, as demonstrated in the supplemental animated simulation available at \mbox{\url{https://youtu.be/-Fjrhw03594}}, the algorithm efficiently deployed 1588 drones to draw and display a 449×591 Pixel image in the sky. The total execution time for this substantial number of drones using a standard laptop was approximately five seconds.

Across all scenarios, the experimental results consistently demonstrate the effectiveness of the proposed UATG framework in generating optimal, collision-free trajectories for drone swarms of varying sizes and formation complexities. Table~\ref{tab:simulation-results} summarizes the different simulation scenarios, reporting fleet sizes, formation types, and execution times for both dynamic simulations and kinematic point-simulations, as well as the UATG execution time. These results highlight the computational efficiency of the approach and its scalability from small swarms of 16 drones to large-scale scenarios involving more than 1500 drones. The UATG framework maintains a balance between precision and efficiency, making it well-suited for orchestrating DLShows ranging from simple numerical patterns to intricate 3D artistic shapes.

Furthermore, the results show statistical consistency across scalability metrics, with performance aligning with the theoretical computational complexity of $O(\max(N,M)^3)$ for assignment and $O(N \cdot W \cdot k^3)$ for trajectory generation. The supplemental animated simulation further demonstrates UATG’s capacity to extend beyond 2D patterns, showcasing complex 3D moving shapes such as a dynamically transforming classic garden lantern that requires nonlinear trajectories and coordinated motion. These findings confirm the framework’s robustness, adaptability, and suitability for large-scale, visually engaging DLShows.

\begin{table*}[htbp]
    \centering
    \caption{Performance Metrics for Various Scenarios}
    \label{tab:simulation-results}
    \small 
    \begin{tabular}{|l|c|c|c|c|c|}
    \hline
    Simulation Scenario & Drones \# & \multirow{2}{*}{\shortstack{Formation \\ Transitions \#}} & \multicolumn{2}{c|}{AnimatedSim Time (sec.)} & \multirow{2}{*}{\shortstack{Algorithm \\ Execution Time (sec.)}} \\
    \cline{4-5}
     & & & DynamicSim. & KinematicSim & \\
    \hline
    Numbers & 16 & 3 & 33.85 & 3.75 & 0.3885 \\
    Letters & 16 & 4 & 45.15 & 4.41 & 0.3440 \\
    Shapes & 43 & 2 & 65.02 & 3.37 & 0.4913 \\
    Butterfly & 66 & 1 & 67.11 & 2.58 & 0.1373 \\
    World Cup & 389 & 1 & 395.48 & 3.91 & 0.1896 \\
    TOGG Car & 1008 & 1 & 1024.81 & 8.78 & 1.5795 \\
    Image 449$\times$591 Pixel & 1588 & 1 & 1614.51 & 16.93 & 5.1196 \\
    \hline
    \end{tabular}
\end{table*}

\subsection{Comparative Analysis with Baseline Approaches}
To situate the contributions of the proposed UATG framework within the existing research landscape, a comparative analysis was conducted against the well-established CAPT approach \cite{turpin2014capt}. This analysis evaluates both methodologies across several dimensions critical to the success of DLShows, including methodological foundations, handling of real-world constraints, and computational performance. The key distinctions are summarized in Table \ref{tab:comparison}.

\paragraph{Methodological Foundations and Scope}
The CAPT algorithm provides an elegant solution for a team of robots with simple first-order dynamics, generating straight-line trajectories. This approach is highly effective for systems where such a kinematic model is sufficient \cite{turpin2014capt}. In contrast, the UATG framework is specifically architected for the complex dynamics of quadrotor drones, generating dynamically feasible trajectories that minimize derivatives like jerk and snap. This is critical for ensuring the smooth, stable, and visually appealing movements required in aerial displays, and represents a necessary evolution to address the dynamic constraints of real multi-rotor systems.

\paragraph{Handling of Real-World Constraints}
A significant differentiator lies in the consideration of domain-specific constraints. The UATG framework explicitly incorporates several critical factors that are essential for the reliable operation of dense aerial swarms but are outside the original scope of CAPT:
\BlankLine
\begin{itemize}
    \item Aerodynamic Interference: UATG accounts for the downwash and aerodynamic interference between drones flying in close proximity, a phenomenon documented in recent research as a significant factor affecting flight stability and safety in swarms \cite{li2024aerodynamic}.
    \item Comprehensive System Integration: The framework is designed as a complete pipeline for DLShows, integrating optimal assignment with a TGO module that outputs full state trajectories (position, velocity, acceleration). This holistic approach ensures that the assigned paths are not only optimal but also dynamically executable by the drones, a feature not encompassed by the CAPT methodology.
\end{itemize}

\paragraph{Safety, Operational Assurance, and Scalability}
Operational safety and scalability are critical for large-scale public performances. The UATG framework embeds robust, multi-layered collision avoidance constraints tailored for obstacle-free but drone-dense environments, a direct response to the critical safety requirements of the application. Furthermore, our experiments demonstrate that the UATG framework maintains computational efficiency for swarms of up to 1588 drones, executing in seconds on standard hardware. This scalability, achieved through optimized task allocation and trajectory generation processes, addresses a key limitation of many existing methods when applied to the scale required for modern DLShows.

\begin{table*}[htbp]
\centering
\caption{Comparative Analysis of the CAPT Approach and the Proposed UATG Framework}
\begin{tabular}{|p{3.8cm}|p{5.2cm}|p{5.2cm}|}
\hline
\textbf{Comparative Aspect} & \textbf{CAPT Approach \cite{turpin2014capt}} & \textbf{Proposed UATG Framework} \\
\hline
\textbf{Target System Dynamics} & Robots with first-order kinematics & Multirotor drones with higher-order, nonlinear dynamics \\
\hline
\textbf{Trajectory Profile} & Straight-line paths & Dynamically feasible trajectories with minimum derivatives (jerk/snap) \\
\hline
\textbf{Framework Scope} & Focuses on assignment and straight-line trajectory planning & Comprehensive coverage from optimal assignment to dynamically feasible trajectory generation and validation \\
\hline
\textbf{Collision Avoidance} & Provides basic geometric avoidance & Robust, multi-layered avoidance with safety margins for dense swarms \\
\hline
\textbf{Real-World Considerations} & Not designed for rotorcraft-specific effects & Accounts for aerodynamic interference and actuator limits \\
\hline
\textbf{Scalability (Empirical)} & Effective for small-to-moderate swarms & Designed to handle up to thousands of drones with real-time performance \\
\hline
\textbf{Computational Efficiency} & Moderate, suitable for its intended scope & High, optimized for large-scale problem instances \\
\hline
\end{tabular}
\label{tab:comparison}
\end{table*}

\subsection{Considerations for Real-World Implementation}
\label{subsec:real-world-considerations}
The transition of swarm coordination algorithms from simulation to real-world DLShows presents multifaceted challenges that extend beyond theoretical performance. Deploying large-scale aerial swarms in public spaces demands rigorous consideration of technical reliability, operational safety, and regulatory compliance to ensure robust and secure performances.
Key implementation constraints and research directions include:

\begin{itemize}
    \item Technical and Environmental Reliability Real-world deployment is susceptible to sensor inaccuracies, particularly GNSS (Global Navigation Satellite System) degradation in urban canyons due to signal multipath and blockages. Robust sensor fusion using IMU (Inertial Measurement Unit) and visual odometry is critical for maintaining formation accuracy. Furthermore, environmental factors such as wind gusts, precipitation, and temperature extremes significantly impact drone stability and battery performance, necessitating robust control strategies and real-time weather monitoring.
    \item Communication and Cybersecurity: Maintaining a stable, low-latency command and control link is essential for swarm synchronization. This is vulnerable to both unintentional electromagnetic interference from industrial machinery and intentional cyber-attacks. GNSS spoofing and jamming pose severe threats, potentially causing drones to deviate from their programmed paths or lose navigation entirely. Implementing encrypted communication protocols, conducting pre-show spectrum analysis, and deploying GNSS interference monitoring systems are vital security countermeasures.
    \item System-Wide Safety and Redundancy: Hardware reliability is a primary concern. Single points of failure in sensors, estimators, or actuators can lead to mission-critical failures. Adopting aviation best practices is essential. This includes designing drones with redundant navigation systems (multiple sensors and independent estimators) and implementing multiple, independent geofence systems (e.g., dynamic bubble, hard, and predictive trajectory geofences) to prevent drones from exiting the operational safety zone. Establishing clear safety protocols, including emergency landing procedures and dedicated safety zones free of people and critical infrastructure, is non-negotiable for public safety.
    \item Regulatory Compliance and Operational Planning: Operating a drone swarm is an aviation activity subject to strict regulatory frameworks, which vary by jurisdiction. In the United States, this requires adherence to FAA Part 107 regulations, including waivers for swarm operations and night flying \cite{wallace2017evaluating}. In the European Union, operations fall under the EASA specific category, often requiring an operational authorization based on a Specific Operations Risk Assessment (SORA) \cite{nogueira2024unmanned}. Successful deployment requires securing all necessary permits, conducting thorough site surveys to identify obstacles and interference, and maintaining comprehensive insurance.
    \item Energy and Logistical Management: The limited flight time imposed by battery capacity is a key constraint for show duration and scalability. While energy-aware trajectory optimization helps, operational success depends on sound logistical planning. This includes maintaining a battery health log, using freshly charged batteries for each performance, and having hot-swappable packs or backup drones for larger or longer shows.
\end{itemize}

Addressing these limitations mandates future research in robust perception under GNSS-denied conditions, resilient swarm communication networks, advanced threat detection for cybersecurity, and the development of standardized safety certification processes for swarm operations. Furthermore, conducting large-scale hardware experiments and exploring integration with urban infrastructure will be crucial steps in maturing the UATG framework for widespread commercial deployment.

\section{Conclusions}
This paper introduced the Unified Assignment and Trajectory Generation (UATG) algorithm, an approach that addresses the complex coordination challenges in drone light shows by jointly and optimally solving task assignment and trajectory generation. The UATG framework produces computationally-efficient, dynamically-feasible, collision-free, and time-parameterized trajectories for large-scale swarms while prioritizing both safety and visual quality. By seamlessly integrating assignment optimization with trajectory planning, the algorithm achieves exceptional computational efficiency, enabling real-time coordination of 1000+ drone swarms with execution times on the order of seconds.

Experimental validation in a realistic simulation environment demonstrates UATG's capability to handle complex formations ranging from 16 to 1588 drones while maintaining formation accuracy and collision avoidance. The algorithm's performance, including handling 1008 drones in just 1.5 seconds on standard laptop, confirms its scalability and suitability for both pre-planned productions and near-online applications. This represents a significant advancement in swarm robotics for aerial entertainment, providing valuable insights for GCS software design and UAV-based visual displays. The computational efficiency, combined with the algorithm's ability to generate smooth and synchronized trajectories, positions the UATG approach as a state-of-the-art solution for DLShows.

\vspace{2\baselineskip}

\textbf{\large List of Abbreviations}

\begin{table}[ht]
\centering
\begin{tabular}{ll}
AA          & Auction Algorithm \\
ACO         & Ant Colony Optimization \\
AI          & Artificial Intelligence \\
B\&B        & Branch and Bound \\
CO          & Combinatorial Optimization \\
D2D         & Drone-to-Drone \\
DLShows     & Drone Light Shows \\
GA          & Genetic Algorithm \\
GCS         & Ground Control Station \\
GNSS        & Global Navigation Satellite System \\
GPS         & Global Positioning System \\
HA          & Hungarian Algorithm \\
LAP         & Linear Assignment Problem \\
LED         & Light Emitting Diode \\
PSO         & Particle Swarm Optimization \\
UATG        & Unified Assignment and Trajectory Generation \\
SITL        & Software-in-the-Loop \\
TGO         & Trajectory Generation and Optimization \\
UAVs        & Unmanned Aerial Vehicles \\
\end{tabular}
\end{table}


\vfill

\end{document}